\documentclass[journal,twoside,web]{ieeecolor}
\usepackage{tmi}
\usepackage{cite}
\usepackage{amsmath,amssymb,amsfonts}
\usepackage{graphicx}
\usepackage{multirow}
\usepackage{booktabs}
\usepackage{textcomp}
\usepackage{algorithm}
\usepackage{algpseudocode}
\usepackage{bm}
\usepackage[outdir=]{epstopdf}

\def\BibTeX{{\rm B\kern-.05em{\sc i\kern-.025em b}\kern-.08em
    T\kern-.1667em\lower.7ex\hbox{E}\kern-.125emX}}

\DeclareUnicodeCharacter{0119}{\l}
\DeclareUnicodeCharacter{0142}{\k{e}}
    
\begin{document}
\title{Tackling the Incomplete Annotation Issue in Universal Lesion Detection Task By Exploratory Training}

\newcommand{\btm}{\textcolor{red}}

\author{
Xiaoyu Bai,
Benteng Ma,
Changyang Li, and
Yong Xia,~\IEEEmembership{Member,~IEEE}
\thanks{This work was supported in part by the National Natural Science Foundation of China under Grants 62171377. ({\em Corresponding authors: Y. Xia}).}
\thanks{X. Bai, B. Ma and Y. Xia are with the National Engineering Laboratory for Integrated Aero-Space-Ground-Ocean Big Data Application Technology, School of Computer Science and Engineering, Northwestern Polytechnical University, Xi’an 710072, China (e-mail: yxia@nwpu.edu.cn).}
\thanks{C. Li is with the Sydney Polytechnic Institute, NSW 2000, Australia (e-mail: chris@ruddergroup.com.au).}
}

\maketitle

\begin{abstract}
Universal lesion detection (ULD) has great value for daily clinical practice as it aims to detect various types of lesions in multiple organs on medical images.   Deep learning methods have shown promising results, but  demanding substantial volumes of annotated data for effective training. 
Nevertheless, annotating medical images is costly and requires specialized knowledge. Additionally, the diverse forms and contrasts of objects in medical images make fully annotation even more challenging, resulting in incomplete annotations. Directly training ULD detectors on such datasets can yield suboptimal results.  Pseudo-label-based methods examine the data in the training set and mine unlabelled objects for retraining, which have shown to be effective to tackle this issue. Presently, top-performing methods rely on a dynamic label-mining mechanism, operating at the mini-batch level. However, the model's performance varies at different iterations, leading to inconsistencies in the quality of the mined labels during training. This inconsistency limits their potential for performance enhancement. Inspired by the observation that deep models learn concepts with increasing complexity, we introduce an innovative exploratory training strategy to assess the reliability of mined lesions over time. Specifically, we introduce a teacher-student detection model as the basis, where the teacher model's predictions are combined with incomplete annotations to train the student model. On top of that, we design a prediction bank to record all high-confidence predictions. Each sample is trained several times, allowing us to get a sequence of records for each sample. If a prediction consistently appears in the record sequence, it is likely to be a true object, otherwise it may just a noise. This serves as a crucial criterion for selecting reliable mined lesions for subsequent retraining. Our experimental results substantiate that the proposed framework surpasses state-of-the-art methods on two medical image datasets, demonstrating its superior performance.
\end{abstract}

\begin{IEEEkeywords}
Missing annotations, DeepLesion, universal lesion detection, memory bank
\end{IEEEkeywords}

\section{Introduction}

Universal lesion detection (ULD) aims to detect various types of lesions in multiple human organs using medical imaging and is of great value in daily clinical practice\cite{arnaud2018fully,yan20183d}. Deep learning methods have shown promise in ULD tasks \cite{yan2018deeplesion,yan2019mulan,zhang2020revisiting}, thanks to their ability to effectively fuse information from multiple images\cite{yan20183d,zhang2020revisiting,yang2021asymmetric,li2022satr} and incorporate auxiliary tasks like segmentation \cite{yan2019mulan,zlocha2019improving}. These advances have led to gradual improvements in overall ULD performance. However, the performance of these methods heavily relies on large-scale training datasets with high-quality annotations. Unfortunately, due to the complex morphology of lesions and the low soft tissue contrast in medical images, it is challenging to gather such datasets. In practice, many lesions may be missed during the annotation process, leading to the issue of incomplete annotations, especially when images contain a large number of lesions.

\begin{figure}[t]
    \centering
    \includegraphics[scale=0.22]{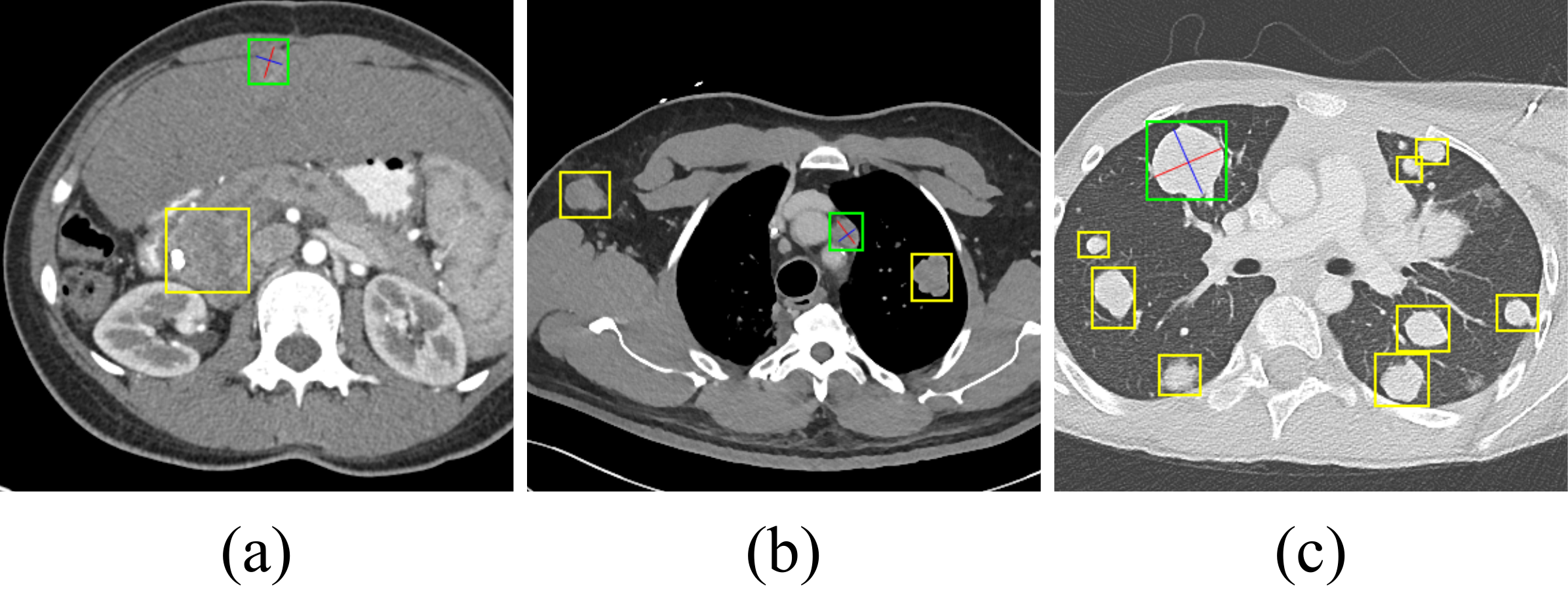}
    \caption{Illustration of incomplete annotation issue. Here are three slices from the DeepLesion dataset, where the bounding boxes generated based on RECIST-style annotations are shown in green, and the bounding boxes of missing annotations of lesions are highlighted in yellow.}    
    \label{fig:1}
\end{figure}

Take the widely-used DeepLesion dataset \cite{yan2018deeplesion} for example, which is a large-scale dataset of lesions from multiple categories and contains RECIST-style annotations \cite{eisenhauer2009new}.
We selected three slices from DeepLesion and displayed them in Fig. \ref{fig:1}, where the bounding boxes generated based on RECIST-style annotations were shown in green, and the bounding boxes of missing annotations of lesions were highlighted in yellow.
It reveals that many lesions were not annotated, particularly in Fig. \ref{fig:1}(c). Indeed, it was reported that DeepLesion have a missing annotation rate of around 50\%\cite{yan2020learning}. 

Training a ULD detector on a dataset with the incomplete annotation issue is challenging, since unlabeled lesions may be misinterpreted as the background, and those false negative samples, in turn, result in degraded performance. Previous study has shown that directly training a ULD detector like MULAN \cite{yan2019mulan} on 
DeepLesion had yielded suboptimal results
\cite{cai2020lesion}. Obviously, addressing the issue of missing annotations is imperative to ensure robust ULD performance.

Although most ULD methods overlook the missing annotation issue, a handful of methods have been proposed to address this issue in the broad field of computer vision. These methods can be generally categorized into three groups.
(1) By assuming that there are no additional objects nearby available annotations, sampling-based methods focus only on those negative samples, which are in close proximity to ground truth annotations\cite{shrivastava2016training, wu2018soft}. In this way, sampling-based methods are able to reduce the number of false negatives, but they ignore many background regions, which may contain valuable discriminative information.
(2) Alternatively, researchers attempted to convert learning from incomplete annotations into learning from complete annotations. Typically, under the selected completely at random (SCAR) assumption, which assume the positive samples are annotated completely at random, the task of object detection with incomplete annotations can be converted to a normal detection task with a re-weighted loss function\cite{yang2020object,zhao2021positive}. This assumption, however, is not always held in practice.
(3) More methods aim to directly mine pseudo labels and use them to guide model training. Pseudo-label-based methods use either a multi-stage training process\cite{cai2020lesion, ying2021semi}, or learning a dynamically changed supervision like co-mining \cite{wang2021co} and loss re-calibration \cite{zhang2020solving,lin2021decoupled}.
Although pseudo-label-based methods have been demonstrated to be most effective in dealing with the incomplete annotation issue, these methods still have major limitations. The multi-stage training involves training an initial detector using incomplete annotations, which results potentially in overfitting and false memorization of unlabeled objects as background and can hardly be recovered in later stages. The dynamic supervision strategy has great potential to overcome this issue, but the pseudo labels generated at each iteration are based on the predictions made by the current model, which means that the quality of these pseudo labels varies over the ongoing training process. Consequently, the pseudo labels generated at different iterations may not possess equal reliability. 

In this paper, we propose the exploratory training for ULD (ET-ULD) method to address the incomplete annotation issue in ULD. It is expected that ET-ULD can overcome the disadvantages of dynamic supervision strategy. Instead of using elaborately designed auxiliary tasks and conservative mining policy to prevent model collapse\cite{lyu2021segmentation,wang2021co}, our ET-ULD deliberately trains a model slowly 
by continuously adding newly-mined pseudo lesions to the training set. To this end, we use a teacher-student model with an exponential moving average (EMA) update strategy as the basis. During training, the high confidence lesions detected by the teacher model are added to the training set to train the student model. Meanwhile, since deep learning models tend to learn simple concepts before gradually memorize all samples\cite{arpit2017closer,kalimeris2019sgd}, we can judge the reliability of each pseudo lesion according to the flip of its label.
Specifically, if a lesion is consistently detected during the training process, it is likely to be a genuine one; otherwise, if a lesion is only detected in the later stage of training or is just detected a few times, it may be a noisy prediction.
Finally, we combine these mined lesions with original annotations and use all of them to retrain the model.

The main contributions of this work are three-fold.
\begin{itemize}    
    \item We propose the ET-ULD method to address the missing annotation issue faced by ULD tasks.
    \item We develop a method to identify highly reliable lesions detected during the training process based on the intrinsic property of deep learning models.
    \item Our ET-ULD method outperforms competing ULD methods on the DeepLesion dataset, setting the new state of the art.
\end{itemize}
\section{Related Work}

\subsection{ULD}
ULD aims to localize various types of lesions all over the body in medical images. This task is challenging, since different lesions may have diverse shapes and sizes and the contrast between lesions and surrounding tissues is usually poor. Deep learning methods have shown promising potential on this task.
Inspired by the observation that radiologists often need multiple slices to locate and diagnose lesions on a single CT slice, most ULD methods use multiple adjacent CT slices as the inputs of a 2D CNN detector \cite{yan20183d,tao2019improving,zlocha2019improving,li2022satr,yang2021asymmetric}. Yan et al. \cite{yan20183d} proposed a novel feature fusion method, in which the features of multiple slices are fused during the intermediate stage of the network. This feature fusion method has been adopted in many studies\cite{cai2020lesion,yan2020learning,wang2019semi}.
Based on a similar idea, Yang et al. \cite{yang2021asymmetric} proposed a 3D context fusion operator called A3D, which effectively integrates contextual information from various 2D slices using asymmetric weighting schemes. 
Li et al. \cite{li2022satr} argued that CNN-based models struggle to capture the global context adequately. To overcome this limitation, they introduced transformer blocks to fuse both intra- and inter-slice features. While most methods employ 2D detectors, Zhang et al. \cite{zhang2020revisiting} demonstrated that using a pseudo 3D network pretrained on the COCO dataset can yield promising results.
Overall, both 2D and pseudo 3D methods have shown good performance in ULD. 2D methods are more popular, since they can benefit from robust 2D models pretrained on large-scale datasets.

All those methods were evaluated on the DeepLesion dataset \cite{yan2018deeplesion}. However, it is worth noting that about 50\% lesions in DeepLesion were not annotated and missing annotations existed in both training set and test set \cite{yan2020learning}. To address the incomplete annotation issue, Cai et al. \cite{cai2020lesion} created a fully-annotated subset of DeepLesion, and developed the Lesion-harvester method to mine unlabeled lesions. In Lesion-harvester, a lesion/non-lesion classifier is trained on the fully annotated subset and used to filter out false predictions made by the detectors trained on the incompletely annotated training set. The remaining predictions are treated as mined lesions and used to train an improved ULD model. Due to involving multiple training stages, Lesion-harvester has significantly increased training time.
Yan et al. \cite{yan2020learning} leveraged multiple fully-annotated single organ datasets to aid the mining of unlabeled lesions. While this approach can enhance detection performance for the organs with additional annotations, the issue of missing annotations remains unresolved for the regions without extra data.

The latest ULD method \cite{lyu2021segmentation} that considers the incomplete annotation issue is based on the co-mining strategy. It consists of a detection branch and a segmentation branch. Each branch can generate lesion detection predictions. But only the predictions, which are either made by a single branch with high confidence or made by both branches consistently, are considered as mined suspicious lesions.
Although this method demonstrates promising detection results, it requires additional pre-and post-processing steps, such as super-pixel generation and mask to bounding box conversion. In theory, incorporating the mined lesions as positive samples for retraining should enhance detection performance. However, in this approach, they have opted to set them as neutral and omitted them from loss computation due to concerns regarding their reliability.

Compared to these methods, our ET-ULD method relies solely on a well-crafted detection model, eliminating the necessity of additional annotations and pre- or post-processing steps. Furthermore, it excels in its capacity to effectively identify highly reliable lesions and incorporate them as positive samples, resulting in a substantial performance enhancement.


\subsection{Pseudo-label-based methods for incomplete annotations}

Nowadays, pseudo-label-based methods are most prevalent for addressing the incomplete annotation issue encountered when training an object detection model. Among these methods, some involve a multi-stage training process. In each stage, the pseudo labels are first corrected using the predictions made in the previous stage and then used to train the model\cite{cai2020lesion, liu2020unbiased}. For instance, Ying et al. \cite{ying2021semi} designed a multi-stage  method for the signet ring cell detection task. This method starts with the training of an initial detector using an incompletely annotated dataset. Subsequently, the detector is employed to identify unlabeled signet ring cells within the training data. These newly discovered cells are then combined with the original annotations to train the next iteration of the model. This process iterates multiple times until no further performance improvement is observed on the validation set. 

This direct multi-stage training-mining strategy has indeed yielded noticeable performance improvements. However, recent researches \cite{cermelli2022modeling,yang2020object} have shed light on a notable concern. Specifically, it has come to attention that a significant number of unlabelled objects are erroneously categorized as part of the background class during the initial model's training phase, and rectifying these errors poses a challenge in subsequent training stages, thereby constraining the overall detection performance.

To mitigate this issue, the mined objects need to be added in the model training process timely. Methods such like co-mining \cite{wang2021co} and loss re-calibration \cite{zhang2020solving,lin2021decoupled} attempt to perform model training and label correction simultaneously in an end-to-end training process.
The co-mining method generates pseudo labels by feeding two augmented views of the same image into a Siamese network and using the predictions from one branch to generate complementary predictions for the other branch. 
Based on the observation that unlabeled foreground objects (false negatives) may generate a larger negative (background) loss than true negative samples, loss re-calibration methods reverse the negative branch of the loss function to the positive branch, if the negative loss is larger than a pre-defined threshold value. 
A drawback of these methods 
is that the pseudo labels generated at each iteration rely on the intermediate model's predictions at that time point. Consequently, the quality of pseudo labels may vary throughout the training process. In other words, the pseudo labels generated at different iterations may not possess equal reliability, but are used indiscriminately, leading to sub-optimal detection performance.

By contrast, our ET-ULD method thoroughly explores the label-mining process and uses an effective method to assess the reliability of the lesions mined at different iterations. By filtering out less reliable lesions, we combine the remaining ones with original annotations for model retraining. In this way, our ET-ULD achieves a substantial performance enhancement.


\begin{figure*}
    \centering
    \includegraphics[scale=0.55]{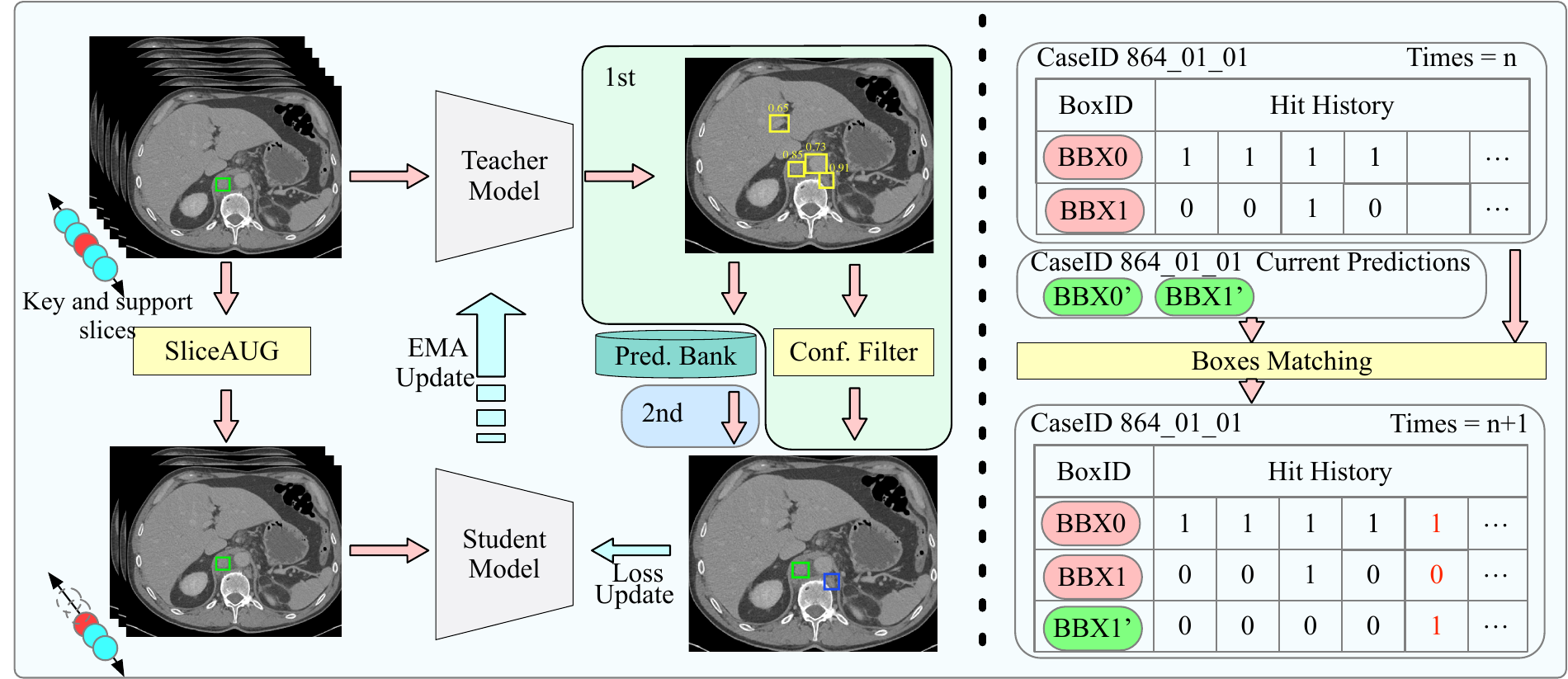}
    \caption{Diagram of our ET-ULD model. The model training process consists of two stages. The exploratory training stage involves generating pseudo labels using the teacher model's predictions and combining them with available ground truth to train the student model. A prediction bank is used to record the pseudo lesion bounding-boxes generated at each iteration. 
    In the second stage, we select the pseudo lesion bounding-boxes with high confidence from the prediction bank combined with the original annotations lesions to retrain the model. 
    The support slices are used to provide the inter-slice context to boost the lesion detection on the key slice. The right part shows an example case in the prediction bank.}
    \label{fig:framework}
\end{figure*}

\section{Method}
\subsection{Problem formulation}
We are provided with a dataset denoted as $\{\bm{X},\bm{Y}\}$, where $\bm{X}$ represents a set of images, and $\bm{Y}$ contains information regarding objects' bounding boxes. The objective of the object detector is to minimize the following loss function:

\begin{equation}
\mathcal{L} = \mathcal{L}_{cls}(\bm{P},\bm{Y}) + \mathcal{L}_{reg}(\bm{P},\bm{Y}).
\end{equation}
Within this function, $\mathcal{L}_{cls}$ is the classification loss, serving as a measure of the agreement between the model's predictions of bounding boxes, denoted as $\bm{P}$, and the actual objects. It's important to note that a bounding box can correspond to either an object or the background class. $\mathcal{L}_{reg}$ represents the regression loss, which quantifies the difference between the predicted bounding boxes and the ground truth bounding boxes. Notably, the computation of $\mathcal{L}_{reg}$ is limited to objects that have been labeled in the dataset, while $\mathcal{L}_{cls}$ encompasses both objects and background regions.  Under the condition of incomplete annotations, the computation of $\mathcal{L}{reg}$ still follows the standard procedure employed in typical object detection tasks. However, this situation significantly impacts the calculation of $\mathcal{L}{cls}$. Hence, our primary focus in this work revolves around addressing the challenges associated with $\mathcal{L}_{cls}$.

Typically, we regard the objects as positive samples and background regions as negative samples. Before delving into the challenges posed by incomplete annotations, let us first establish the foundation by examining the learning process under fully annotations. Given  a fully annotated dataset, the term $\mathcal{L}_{cls}$ can be viewed as a positive-negative (PN) classification problem. To formalize this, consider a joint distribution $p(\bm{x},y)$ of $(\bm{X},Y)$, where $\bm{X} \in \mathbb{R}^d$ represents the data and $Y \in\{-1,1\}$ serves as the label. Our goal is to develop a model $f$ capable of effectively distinguishing positive and negative samples. Theoretically, this is achieved by minimizing the expected risk:

\begin{equation}
R = \mathbb{E}_p(l(f(\bm{x}),y)),
\label{func:1}
\end{equation}
where $l$ is the loss function which measures the deviation between the output of the model $f$  and the label. 
Alternatively, we can consider the data distribution $p(\bm{x})$ as a weighted combination of the positive distribution $p_p(\bm{x})$ and the negative distribution $p_n(\bm{x})$, i.e., $p(\bm{x}) =\pi_pp_p(\bm{x})+\pi_np_n(\bm{x})$, with $\pi_p$ and $\pi_n$ representing the class priors for the positive and negative classes, respectively. We can rewrite equation (\ref{func:1}) as:
\begin{equation}
  R = \pi_p\mathbb{E}_p(l(f(\bm{x}),1))+\pi_n\mathbb{E}_n(l(f(\bm{x}),-1)).
\end{equation}
Since the actual data distribution is unavailable, in practice, we measure the performance on a known training set $\bm{X}_{train}=\{\bm{x}_i\}_{i=1}^{N}\overset{i.i.d.}{\sim}p(\bm{x})$, giving us the empirical risk:

\begin{equation}
R_e = \pi_pR^+(\bm{X}_p) + \pi_nR^-(\bm{X}_n),
\label{func:er}
\end{equation}
where $R^+(\bm{X}_p) = 1/M\sum_{i=1}^{M}H(f(\bm{x}_i^p),1)$, $R^-(\bm{X}_n) = 1/(N-M)\sum_{i=1}^{N-M}H(f(\bm{x}_i^n),-1)$,
 $M$ is the number of positive instances, $\pi_p = \frac{M}{N}$ and $\pi_n=1-\pi_p$ are class priors, $H(\cdot,\cdot)$ denotes the cross-entropy loss function. 

If the annotations are incomplete, the presence of unannotated positive samples can introduce complications into the learning process. It's common for human annotators to focus on labeling positive samples while leaving others as negative samples. As a result, unlabeled positive samples are mistakenly treated as negative samples, leading to potentially detrimental effects on the computation of $\pi_nR^-(\bm{X}_n)$.

In this scenario, the dataset is referred to as positive and unlabeled (PU) data. Here, the labeled data are all  positive samples, while the unlabeled data comprises a combination of positive samples without annotations and negative samples. To learn from PU data,
we need to make assumptions of the data  distribution\cite{kato2018learning}.  As mentioned in introduction, one choice is use the SCAR assumption, which states that all labeled instances, regardless of their characteristics, are randomly and independently selected from the positive distribution $p_p(\bm{x})$, i.e. $p_l(\bm{x}) = p_p(\bm{x})$.  Also, the unlabeled instances are randomly selected from the $p(\bm{x})$, thus $p_u(x) = p(\bm{x})$. Therefore, we have $\pi_np_n(\bm{x}) = p_u(\bm{x}) - \pi_pp_l(\bm{x})$, $\pi_n = 1 -\pi_p$.  The  risk under the SCAR assumption is:
\begin{equation}
\begin{split}
  R_{scar} = &\pi_p\mathbb{E}_l(l(f(\bm{x}),1))+\mathbb{E}_u(l(f(\bm{x}),-1))\\
  &-\pi_p\mathbb{E}_l(l(f(\bm{x}),-1)).
 \end{split}
\end{equation}
The corresponding empirical risk is
\begin{equation}
  R_{escar} = \pi_pR_l^+(\bm{X}_l)+R_u^-(\bm{X}_u) -\pi_pR_l^-(\bm{X}_l).
  \label{function:4}
\end{equation}
In comparison to the empirical risk (\ref{func:er}) of a PN dataset, the empirical risk under the SCAR assumption incorporates an additional corrective term to accommodate unlabelled positive samples, as expressed by $\pi_nR^-(\bm{X}_n) = R_u^-(\bm{X}_u) -\pi_pR_l^-(\bm{X}_l)$. Although this appears to address the issue, the SCAR assumption doesn't always hold in practical scenarios. For instance, based on clinical requirements, radiologists may choose to label only a specific class of tumors or focus on measuring large tumors, leaving other types of lesions unlabelled. In such cases, SCAR-based methods are not suitable.

Another choice is to make the class separability assumption, which assumes (1) the foreground and background distributions are separable, and (2) similar instances are likely to have the same label.

Formally, the first assumption tell us there exists an indicator function $g$, for any instance $\bm{x}$, it holds
\begin{equation}
\begin{split}
        g(\bm{x}) &	\geq \gamma,  y=1 \\
        g(\bm{x}) &< \gamma,   y=0.
\end{split}
\end{equation}
The second assumption means for two similar instances $\bm{x}$ and $\bm{x}'$, the $p(y\vert \bm{x})$ is also similar to $p(y' \vert \bm{x}')$. 
Under these two assumptions, suppose we have already be given a perfect indicator function $g$, we can derive the correct empirical risk:
\begin{equation}
\begin{split}
      R_{e} &= \pi_lR_l^+(\bm{X}_l)+\pi_uR_u(\bm{X}_u \vert g(\bm{X}_u))\\
      &=\pi_lR_l^+(\bm{X}_l)+\pi_u^+R_u^+(\bm{X}_u\vert g(\bm{X}_u)\geq \gamma)\\
      &+\pi_u^-R_u^-(\bm{X}_u\vert g(\bm{X}_u)<\gamma).
\end{split}
\label{func:er_g}
\end{equation}

Apparently, the perfect indicator function $g$ is  our desired model $f$, yet remains inaccessible. In practical applications, both sampling-based methods and pseudo-label-based methods depend on approximating $g$ to make the learning process feasible.

\textbf{Sampling-based methods} use distance related rules to filter out regions far from the annotated ones. Using $g_{dist}(x)$ to represent the distance between the unlabeled regions and the nearest labeled object,  the objective function of the sampling based methods can be formed as:
\begin{equation}
\begin{split}
      \mathcal{L}_{sampling} &= \pi_lR_l^+(\bm{X}_l)+\pi_u^-R_u^-(\bm{X}_u\vert g_{dist}(\bm{X}_u)<\gamma).
\end{split}
\label{func:e_sampling}
\end{equation}

\textbf{Pseudo-label-based methods} commonly involve an iterative learning process. Here, we use a unified representation $f^{i}$ to denote the $i_{th}$ stage model in multi-stage methods and $i_{th}$ mini-batch iterations for dynamic approaches. The previous model $f^{i-1}$ is used to classify the unlabeled data, and the predictions are then utilized to update the current model $f^i$. The objective function is
\begin{equation}
\begin{split}
       \mathcal{L}_{pseudo}^i &= \pi_pR_l^+(\bm{X}_l)+\pi_u^+R_u^+(\bm{X}_u\vert f^{i-1}(\bm{X}_u)\geq\gamma)\\
      &+\pi_u^+R_u^-(\bm{X}_u\vert f^{i-1}(\bm{X}_u)<\gamma).
\end{split}
\label{func:e_re}
\end{equation}


From the comparison above, it is clear that sampling-based methods tend to overlook the unlabeled regions, which limits their ability to accurately approximate the empirical risk. On the other hand, if the mined labels are reliable, pseudo-label methods have the potential to progressively converge on the correct empirical risk. However, in the latest pseudo-label-based methods like co-mining and loss re-calibration, the model's ability fluctuates as it undergoes training, resulting in varying reliability of the mined objects across different iterations. Utilizing these objects indiscriminately would result in sub-optimal performance.
Hence, there arises a need for a mechanism to assess the mined objects in a more granular manner, ideally on a per-iteration basis. As we will elucidate, our method introduces a robust solution to evaluate the reliability of these mined objects at different iterations.

\subsection{Our model}
Our method employs a teacher-student model with an exponential moving average (EMA) update strategy as its basis. The model structure is depicted in Figure \ref{fig:framework}. During training, the teacher model is utilized to infer the input data. Predictions made by the teacher model that pass a confidence filter are then combined with available incomplete labels to train the student model. Meanwhile, a prediction bank is used to record all the label mining histories. In comparison to other approaches, the EMA teacher-student model has several notable advantages. It does not require the design of synergistic tasks, as  previous multi-task methods do, and the EMA update strategy has a model ensemble effect, resulting in more reliable results compared to the Siamese model.
The initial training of the student model is done using only the available annotations in the incompletely labeled dataset $\{\bm{x}_i,\bm{y}_i^l,\bm{y}_i^u\}_{i=1}^N$, where $\bm{x}_i$ represents the image, $\bm{y}_i^l$ denotes available annotations, $\bm{y}_i^u$ are missing annotations. During the training process, the teacher model gradually gains the capacity and starts generate predictions  $\{\bm{p}_i^{\rm{T}} \}_{i=1}^{N_b}$ for the batch inputs $\{\bm{x}_i\}_{i=1}^{N_b}$. These predictions are then compared to the available ground truth $\{\bm{y}_i^l\}_{i=1}^{N_b}$using the intersection over union (IOU) values. The predictions that have an IOU above a certain threshold with any ground truth are removed using a process called GT NMS, which is a Non-Maximum Suppression (NMS) operation on the predictions and ground truth. The remaining predictions are used to train the student model, but only if their confidence scores are above a threshold $\tau$.
\subsection{Exploratory training}
During the training process, the teacher model gradually adds new mined objects to train the student model, a process we can think of as pushing the model's classification boundary from objects with annotations to other possible objects. Previous studies have shown that deep learning models tend to learn simple concepts before gradually memorize all samples\cite{arpit2017closer,kalimeris2019sgd}. In our case, unlabeled samples closely resemble annotated lesions, representing what we can term as 'simple concepts'. Therefore, within the feature space, samples situated in close proximity to existing annotations are likely to be the first ones to traverse the classification boundary and be recognized as new positive samples. According to the class separability assumption, these samples are likely to be true lesions, so at the beginning, the pseudo-labels generated by the teacher model are expected to exhibit a high degree of reliability, representing easily identifiable positive samples. As these easy samples are mined, the decision boundary is pushed further from the available ground truth, moving to regions where sample labels are more uncertain. These samples are a combination of hard positive and hard negative samples. If the training process continues, these uncertain samples will join in to train the student model, leaving only easy negative samples still classified as negative.
We refer to this process as exploratory training because it involves exploring the potential of flipping an unlabeled sample as positive. We design a prediction bank to record the process of exploratory training. Specifically, we initialize the prediction bank for each case in the training set using the incomplete annotations. In each mini-batch, for a specific case, the teacher model will generate a set of predictions.  We use these predictions to both train student model and update the prediction bank. 
The predictions contain three types of bounding boxes: the first represents the same objects as the ground truth, so we use the GT NMS operation to remove these predictions and only keep the ground truth. The second type are predictions that have occurred in the prediction bank in previous iterations, so we match their bounding boxes using an IOU filter. The bounding box locations may slightly change between the prediction bank and the current iteration's prediction, we simply update them using the average. The third type refers to bounding boxes that have never existed in the prediction bank before, so they are added. Also, it is possible that one bounding box appear in the previous prediction bank, but missing in the current mini-batch's prediction, we keep record this box as it may appear again in the following iterations.
To make the prediction bank traceable, we only record bounding boxes with a prediction score higher than a threshold value $\theta$. We further simplify this record as a binary hit history. If a bounding box occurs and its score is higher than $\theta$, we record a 1 in the hit history; otherwise, we record a 0. This update rule is summarized in Algorithm \ref{alg:pre}.

\begin{algorithm}
\caption{Prediction bank update rule}\label{alg:pre}
\begin{algorithmic}[1]
\Require occurrence number $n$, input $x$, incomplete annotations $y^l$, prediction bank $P$, a set of predictions $P^t$
\Ensure updated prediction bank $P$
\If{$n$==0}
    \State $P = y^l$
\Else
\State $P^t = P^t[score(P^t)>\theta]$ \Comment{Filter our low score boxes}
\State  $P^t = P^t[\rm{GTNMS}(P^t,y^l)]$ \Comment{Filter out GTs}
\State $m=$len$(P^t)$,$k=$len$(P)$ \Comment{Number of boxes}
\For {$BOX_i$ in $P$}  
\For {$BOX_j$ in $P^t$} 
\If{IOU$(BOX_j,BOX_i)> \gamma$}
    \State  $BOX_i$ = average$(BOX_i$,$BOX_j)$
    \State $BOX_i\_hithistory$.append(1)
    \State $flag = 1$
\ElsIf{$flag==0$ and $j==k$}
    \State $BOX_i\_hithistory$.append(0)
\EndIf
\EndFor
\EndFor
\For {$BOX_j$ in $P^t$}
\For {$BOX_i$ in $P$} 
\If{IOU$(BOX_j,BOX_i)< \gamma$ and $i==m$}
    \State  add $BOX_j$ in $P_i$
    \State $BOX_j\_hithistory$.append(0)
\EndIf

\EndFor
\EndFor
\EndIf
\end{algorithmic}
\end{algorithm}
As previously discussed, the accumulation of noise during exploratory training can eventually lead to model collapse. This brings forth a crucial question: when should we conclude the exploratory training?
In cases where a fully annotated validation set is available, the decision regarding the total number of iterations for the exploratory training can be determined by the model's performance on this set.
Alternatively, the count of mined objects within the prediction bank can also serve as a robust indicator for terminating the exploration training process. This is because, as the exploration training progresses into its later iterations, an increasing amount of noise gets introduced into the prediction bank. Consequently, the total number of mined objects undergoes a significant surge, eventually surpassing the potential number of true objects. It is advisable to stop the training when this occurs.
\subsection{Selective retraining, data augmentation and inference}
We then select high reliable pseudo labels from the prediction bank for retraining. As previously discussed, boxes that consistently appear in the prediction bank are very likely to be unlabeled objects. On the other hand, if a box only appears in the later period or just appear a few times, it may be a noisy prediction. By using the summation of the hit history for each box as the confidence measure, we can select high reliable pseudo labels.

For the task of ULD, similar to previous studies \cite{cai2020lesion,lyu2021segmentation}, we use consecutive slices as support slices to provide context information for the key slice. While a key slice in the center of the CT volume can have context information from both sides, slices near the boundary of the volume will have less information from one side. This inconsistency can degrade the model's performance on the marginal slices. To address this issue, we introduce a SliceDropout augmentation method. For key slices in the center of the CT volume, we randomly drop some support slices from one side to mimic the situation for boundary slices. We also employ random resize and random box jitter for both the teacher and student models in the training process. At inference time, we input the 3D volume slice by slice into the model and obtain a sequence of 2D predictions for each slice in the input volume. We then use the Kalman filter, as in the lesion harvester \cite{cai2020lesion}, to trace the 2D predictions on each slice to 3D lesion bounding boxes.

In summary, the objective function for exploratory training is expressed as follows:
\begin{equation}
\begin{split}
       \mathcal{L}_{explor} &= \pi_pR_l^+(\bm{X}_l)+\pi_u^+R_u^+(\bm{X}_u\vert f_{ema}(\bm{X}_u)\geq\tau)\\
      &+\pi_u^-R_u^-(\bm{X}_u\vert f_{ema}(\bm{X}_u)<\tau).
\end{split}
\label{func:e_explor}
\end{equation}
Simultaneously, we perform an update on the prediction bank denoted as $p$ using the following procedure:
\begin{equation}
\begin{split}
      p.update(f_{ema}(\bm{X}_u)\geq \theta).
\end{split}
\label{func:e_update}
\end{equation}
Finally, the model is retrained using the subsequent objective function:
\begin{equation}
\begin{split}
       \mathcal{L}_{retrain} &= \pi_pR_l^+(\bm{X}_l)+\pi_u^+R_u^+(\bm{X}_u\vert p.count(\bm{X}_u)\geq\epsilon)\\
      &+\pi_u^-R_u^-(\bm{X}_u\vert p.count(\bm{X}_u)<\epsilon).
\end{split}
\label{func:e_retrain}
\end{equation}
Here, in the context of $p.count$, it denotes the cumulative sum of the hit history for each bounding box, and $\epsilon$ represents a predefined threshold value.

\section{Experiments}
\subsection{Dataset and Metrics}

The DeepLesion dataset \cite{yan2018deeplesion} is a commonly used CT dataset for universal lesion detection, but it has a high rate of missing annotations, with approximately 50\% of the lesions unlabeled. In order to address this issue, Cai et al. \cite{cai2020lesion} fully annotated and released 1915 subvolumes from the DeepLesion dataset, including 844 subvolumes from the original training set (744 for additional training and 100 for validation) and 1,071 subvolumes from the original test set. In this study, we use this same dataset, using the 100 fully annotated subvolumes for validation and the rest of the original training set for training. The 1,071 fully annotated subvolumes are used for test. We report the average precision and detection recalls of our method at operating points with false positive (FP) rates ranging from 0.125 to 8 per volume, using the P3D IoU evaluation metric \cite{cai2020lesion}.

In addition, we tested our method on the LiTS challenge dataset \cite{bilic2023liver}. This dataset comprises contrast-enhanced abdominal CT scans, along with liver and liver tumor masks. Since annotations for the test set are not available, we used only the 131 training sets, splitting them from No.0 to No.96 for training and the remaining for validation. To simulate incomplete annotations, we generated bounding boxes for each tumor using only the center slice of the tumor mask. Therefore, if multiple tumors appear in the center slice of a tumor, they are likely un-annotated because this slice is not the center of their tumor masks.

  \begin{figure}
    \centering
\includegraphics[scale=0.65]{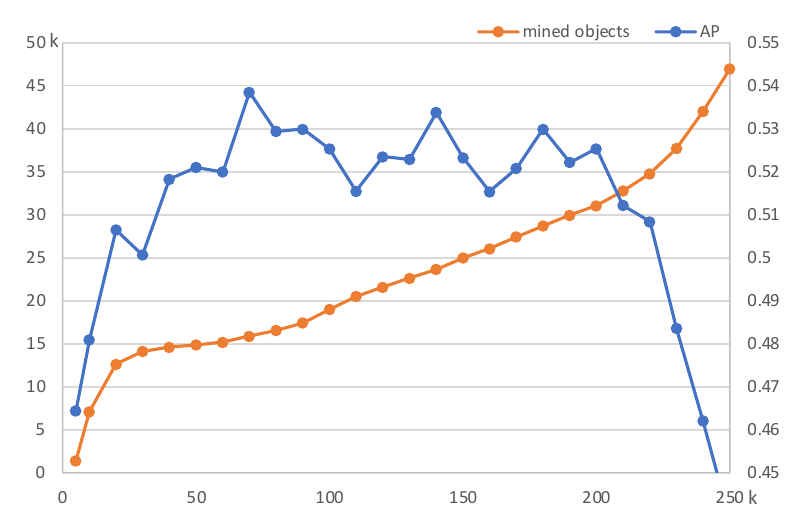}
    \caption{The number of mined objects and the average precision (AP) on the DeepLesion validation set during the  Exploratory training. }
    \label{fig:exp}
\end{figure}

  \begin{figure}
    \centering
\includegraphics[scale=0.4]{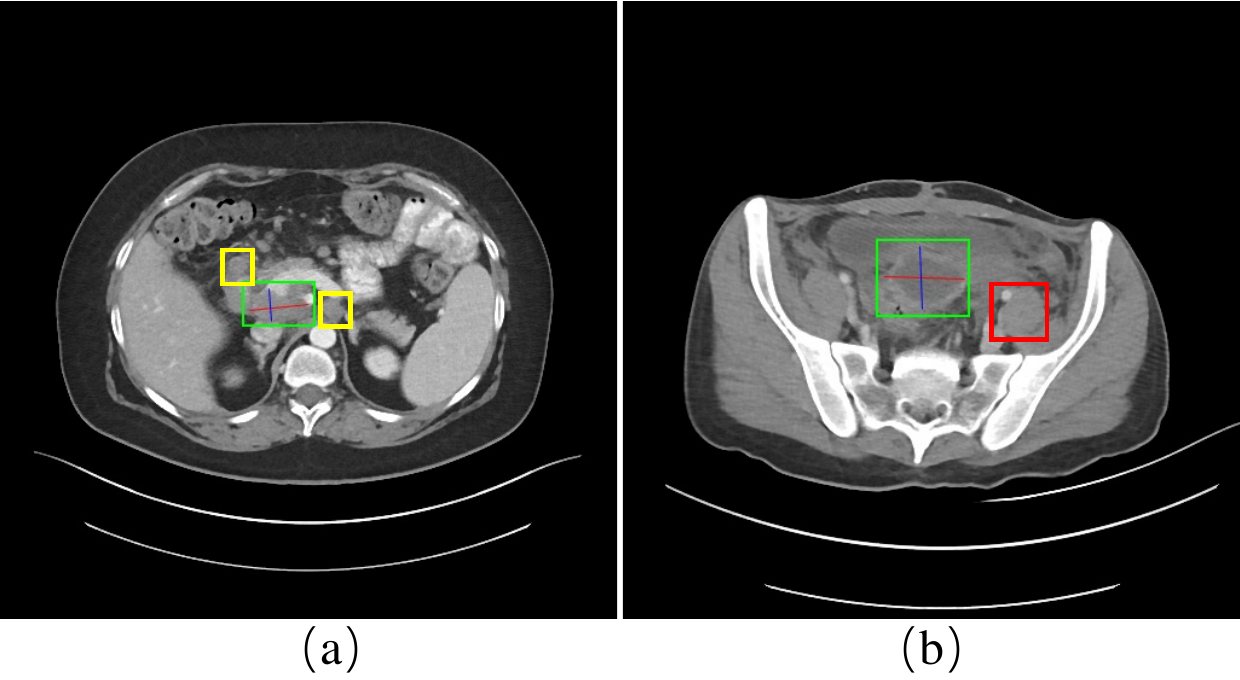}
    \caption{The mined suspicious lesions  are recorded in the prediction bank. Two lesions, shown in yellow bounding boxes in (a), consistently appear in the prediction bank. However, the object shown as a red bounding box in (b) is only added to the prediction bank during late training iterations. The yellow bounding boxes contain true unlabeled lesions, while the red bounding box represents muscle.}
    \label{fig:example}
\end{figure}

 \begin{table*}[]
\centering
\begin{tabular}{@{}l|ccc|cccccccc|c@{}}
\toprule
 \multicolumn{1}{c|}{\multirow{3}{*}{Method}}&  &  &  & \multicolumn{8}{c|}{Sensitivity (\%) at different FPs per subvolume} &  \\ \cmidrule(lr){5-12}
 & \multirow{-2}{*}{R} & \multirow{-2}{*}{P+} & \multirow{-2}{*}{P-} & 0.125 & 0.25 & 0.5 & 1 & 2 & 4 & 8 & AVG & \multirow{-2}{*}{AP (\%)} \\
  \midrule
 \multicolumn{12}{c}{LiTS Results} \\
 \midrule
 Faster-RCNN\cite{yan2019mulan} &  \checkmark &  &  & 25.00 & 34.16 & 39.16 & 49.16 & 55.83 & 67.50 &70.00 & 48.68 & 52.9 \\
Multi-stage Retraining \cite{cai2020lesion}& \checkmark &  \checkmark &  &31.66 & 40.83 & 50.83 & 58.33 &59.16 & 64.16 &65.83 & 52.97 & 55.2\\
 Loss Re-calibration\cite{zhang2020solving}& \checkmark &\checkmark  &  & 37.50 & 45.83 & \textbf{55.83} & \textbf{63.33} & 65.83 & 68.33 &68.33 & 57.85 & 61.3\\
 
Ours &  \checkmark & \checkmark  &  & \textbf{44.16} & \textbf{48.33} & 53.33 & \textbf{63.33} & \textbf{68.33} & \textbf{75.00} &\textbf{78.33} & \textbf{61.54} & \textbf{66.2} \\
 \midrule
 \multicolumn{12}{c}{DeepLesion Results} \\
 \midrule
MULAN\cite{yan2019mulan} &  \checkmark &  &  & 11.43 & 18.69 & 26.98 & 38.99 & 50.15 & 60.38 & 69.71 & 39.47 & 41.8 \\
Lesion   Harvester \cite{cai2020lesion} & { \checkmark} &  &  & 11.92 & 18.42 & 27.54 & 38.91 & 50.15 & \textbf{60.76} & \textbf{69.82} & 39.64 & 43.0 \\
Mask R-CNN \cite{he2017mask} & { \checkmark} &  &  & 13.64 & 19.92 & 27.48 & 36.73 & 47.01 & 57.41 & 66.38 & 38.36 & 41.4 \\
Segmentation-Assisted \cite{lyu2021segmentation} & { \checkmark} &  &  & 14.84 & 20.08 & 27.93 & 37.14 & 47.59 & 57.65 & 66.47 & 38.81 & 42.1 \\
EMA Ensembled Model & { \checkmark} &  &  & \textbf{19.24} & \textbf{25.05} & \textbf{33.04} & \textbf{41.39} & \textbf{50.61} & 58.07 & 65.77 & \textbf{41.88} & \textbf{45.8} \\
\midrule
Wang \cite{wang2019semi} & \checkmark & \checkmark &  & 15.88 & 20.98 & 29.41 & 38.36 & 47.89 & 58.14 & 67.89 & 39.79 & 43.2 \\
Loss re-calibration\cite{zhang2020solving} & \checkmark & \checkmark &  & 15.68 & 23.38 & 31.62 & 40.62 & 48.40 & 56.78 & 64.20 & 40.10 & 42.5 \\
Lesion   Harvester \cite{cai2020lesion} & \checkmark & \checkmark &  & 13.41 & 19.16 & 27.34 & 37.54 & 49.33 & 60.52 & 70.18 & 39.64 & 43.6 \\
Segmentation-Assisted \cite{lyu2021segmentation} & \checkmark & \checkmark &  & 20.01 & 25.65 & 33.44 & 42.81 & 52.39 & 60.69 & 68.79 & 43.39 & 47.5 \\

Ours & \checkmark & \checkmark &  & \textbf{20.76} & \textbf{27.87}  & \textbf{38.09} & \textbf{48.65} & \textbf{57.90} & \textbf{66.66} & \textbf{73.70} & \textbf{47.66} & \textbf{52.6} \\
\midrule
lesion   Havester \cite{cai2020lesion}& \checkmark & \checkmark & \checkmark & 19.86 & 27.11 & 36.21 & 46.82 & 56.89 & 66.82 & 74.73 & 46.92 & 51.9 \\
Segmentation-Assisted \cite{lyu2021segmentation}& \checkmark & \checkmark & \checkmark & \textbf{30.89} & \textbf{37.28} & \textbf{43.94} & 51.05 & 57.41 & 63.68 & 69.81 & 50.58 & 54.7 \\
Ours & \checkmark & \checkmark & \checkmark & 26.57 & 33.56 & 42.31 & \textbf{51.15} & \textbf{60.13} & \textbf{69.79} & \textbf{77.13} & \textbf{51.52} & \textbf{57.1} \\ \bottomrule
\end{tabular}
\caption{Sensitivity at various FPs per image on the completely annotated testing set of DeepLesion and LiTS, we denote the official annotated lesions, mined suspicious lesions and mined hard negative samples as R, P+ and P- respectively.}
\label{tab:comparison}
\end{table*}
\subsection{Implementation Details}
Similar to other methods, we employ Faster-RCNN with a truncated DenseNet-121 pre-trained on ImageNet as its backbone. To enhance localization ability, we utilize a double head design for the ROI head, as suggested in \cite{wu2020rethinking}.
 Each input contains 9 contiguous CT slices, as in previous studies \cite{cai2020lesion,lyu2021segmentation}. The CT slices are interpolated to 2mm in the z-axis and 0.8mm in the xy-plane, and the 12-bit CT intensity range is rescaled to [0,255] using a single windowing (-1024–3071 HU). The random resize ratio for the student model is set to [0.8,1.2], and the IOU threshold for the GT NMS operation is 0.7. The chance of sliceDropout is set to 0.5. 
 The confidence filter threshold $\tau$  is set to 0.9, and the  prediction bank record threshold $\theta$ is set to 0.85.
We set the learning rate as 0.02 and the momentum value for the EMA update is set to 0.999. We use the Stochastic Gradient Descent (SGD) optimizer to train the model. For the Exploratory training, we train the model for 250k iterations.

\subsection{Exploratory training results}

Figure \ref{fig:exp} illustrates the exploratory training process on the DeepLesion dataset. The number of suspicious lesions mined during the current training iterations is approximated by counting the occurrences of "1" in the latest column of the prediction bank. Additionally, we monitored the model's average precision (AP) on the validation set. As shown, the number of mined objects gradually increased as training progressed. Initially, the model efficiently mined easy positive samples, resulting in improved performance. However, as training continued, the model's performance fluctuated, and the rate of mining objects slowed down. This indicates a shifting decision boundary towards more uncertain and low-density regions. Eventually, the boundary will cross over uncertain areas, causing more negative samples to be incorrectly identified as positive, leading to model collapse. Figure \ref{fig:example} presents the  visualization of the mined objects at different iterations of training. It is evident that the objects mined at later training iterations are unreliable. Based on the results from the validation set, we selected the prediction bank from 150k iterations for retraining, as later iterations seemed to introduce more noise. For selective retraining, since each sample in the prediction bank occurs approximately 50 times, we tested the threshold values of 10, 20, and 30 for $\epsilon$, and found that a threshold of 20 yielded the best overall performance.
\subsection{Main Results}
 \begin{figure*}
    \centering
\includegraphics[scale=0.23]{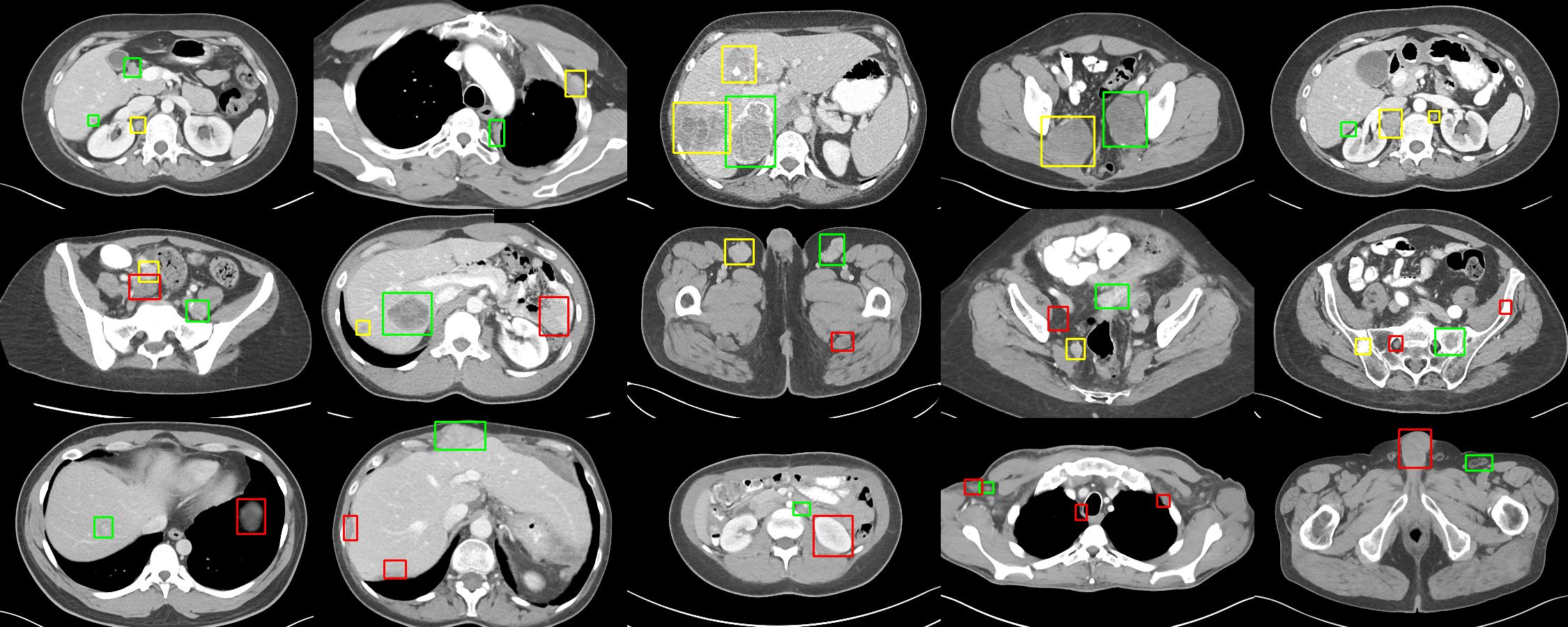}
    \caption{More examples of the mined record in the prediction bank. The green bounding boxes correspond to the existing incomplete annotations, while the yellow bounding boxes represent mined objects that have appeared more than 20 times in total. Conversely, the red boxes denote objects that have appeared fewer than 5 times.}
    \label{fig:visual}
\end{figure*}
Table \ref{tab:comparison} compares the performance of our proposed method with previous state-of-the-art methods for ULD on the fully annotated DeepLesion test set. MULAN \cite{yan2019mulan} is one of the highest-performing methods on the original partially labeled test set. Wang et al. \cite{wang2019semi}, Lesion Harvester \cite{cai2020lesion}, and Segmentation-Assisted \cite{lyu2021segmentation} are three recent methods considered for mining unlabeled lesions. In the Segmentation-Assisted paper, the performance of the Mask-RCNN method \cite{he2017mask} is also reported and is included in the table. 
Additionally, we implemented the loss re-calibration method \cite{zhang2020solving} by reversing the background loss to foreground loss if the background loss is larger than a threshold. We set the threshold to 0.95, as we found that using a lower value caused the model to diverge easily.
 We first present the results trained only using the original annotations. As observed, the EMA ensembled teacher model achieves the overall best performance, demonstrating the useful ensemble effect of the EMA  model.

Regarding the lesion mining results, our method surpasses all other methods by a large margin. In particular, it outperforms the state-of-the-art Segmentation-Assisted method \cite{lyu2021segmentation} by 4.2\% in average sensitivity and 5.1\% in average precision, proving the superior lesion mining capacity of our method. To ensure a thorough comparison, we also applied our trained detector to fully annotated training subvolumes and selected hard negative samples for retraining, following the same approach outlined in \cite{lyu2021segmentation}. Our method exhibits higher recall while maintaining reasonable sensitivity and achieves the best overall metric results.

We also present the results obtained from the LiTS dataset. The Faster-RCNN, trained on the original incomplete annotations, serves as the baseline. The multi-stage retraining demonstrates an increase in the AP from 52.9\% to 55.2\%. However, it falls significantly behind the performance achieved by the loss re-calibration method, highlighting the limitations of the straightforward multi-stage approach. In contrast, our method surpasses the loss re-calibration method by a large margin. When comparing the results on both test sets, the loss re-calibration method exhibits better performance on the LiTS dataset, but does not show such a significant improvement on the much harder DeepLesion dataset. On the contrary, our method significantly boosts performance on both datasets.

\subsection{Ablation Studies}

\begin{figure}
    \centering
\includegraphics[scale=0.52]{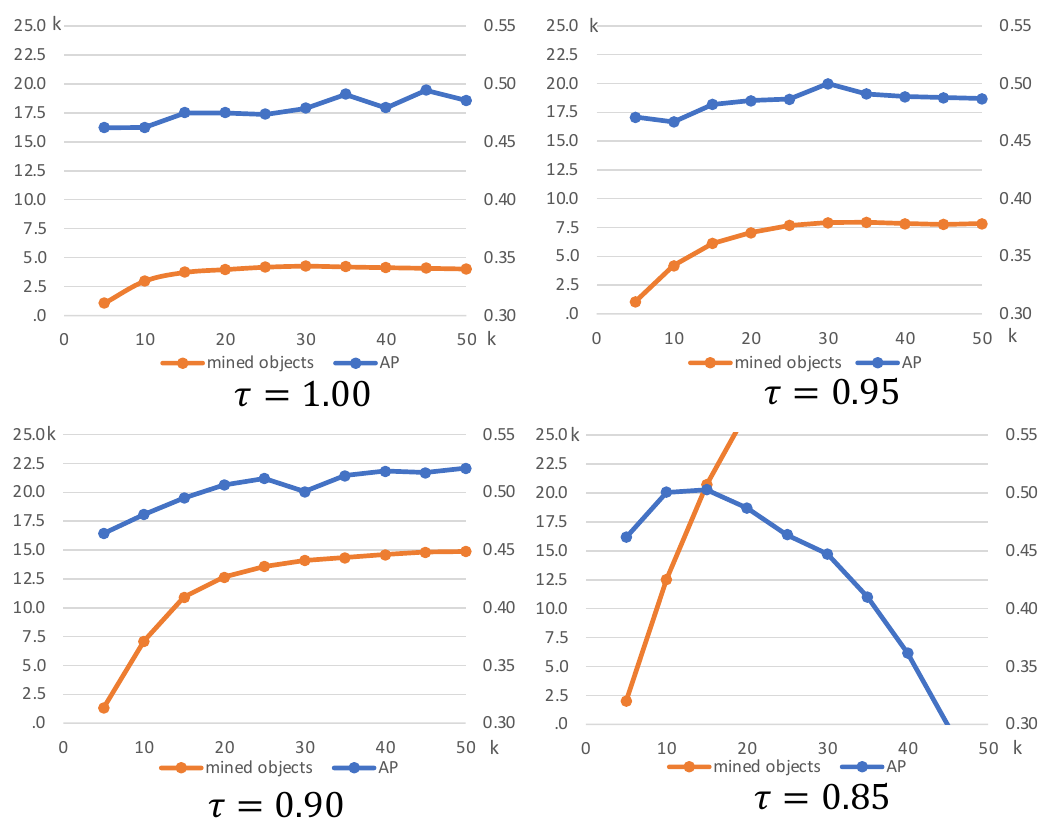}
    \caption{The number of mined lesions on the DeepLesion training set and AP on the validation set under different $\tau$ value. }
    \label{fig:mine_thres}
\end{figure}

\begin{table}
\centering
\begin{tabular}{ccc}
\toprule
Components  & Average Sensitivity (\%)& AP (\%) \\
\midrule
Base model &45.55 & 49.8 \\
+ double head  &45.83&50.2 \\
+ sliceAUG  &46.11&50.9 \\
+ selective retraining &47.66  &52.6 \\
\bottomrule
\end{tabular}
\caption{ Ablations on model components. }
\label{tab:components}
\end{table}

\begin{table}
\centering
\begin{tabular}{ccc}
\toprule
Strategy  & Average Sensitivity (\%) & AP (\%) \\
\midrule
Adding & 47.66 & 52.6\\
Ignoring & 46.13 & 51.0 \\
\bottomrule
\end{tabular}
\caption{Comparison of different  strategies for using the mined lesions.}
\label{tab:stra}
\end{table}

\subsubsection{Effects of each component}
We begin by verifying the contribution of each component to our ET-ULD model. As shown in Table \ref{tab:components}, the base model indicates the best result from the explore training only (based on the validation set's result), which yields a respectable result of 49.8\% AP and an average sensitivity of 45.55\%. Incorporating the double head design slightly enhances the model's performance by approximately 0.4\%. Furthermore, employing sliceAUG leads to an additional increase of 0.5\% in AP. Finally, through the utilization of selective retraining, the AP can be boosted by 1.7\%, demonstrating the effectiveness of our proposed method.

\subsubsection{Effects of The Confidence Filter's Threshold}
In the explore training step, we utilize a fixed threshold value $\tau$ to filter the predictions of the teacher model and generate pseudo labels for training the student model. By extracting data from the prediction bank, we can visualize the mining dynamics across different $\tau$ values. In Figure \ref{fig:mine_thres}, we first present the results when $\tau$ is set to 1, indicating that no pseudo objects are utilized in training the student model. As observed, the number of mined objects recorded in the prediction bank increases slowly, reaching a total of less than 5k.
By decreasing the $\tau$ value to 0.95, high-confidence pseudo labels begin to supervise the training of the student model. After 50k iterations, the number of mined lesions is approximately 7.5k, and the average precision (AP) slightly improves compared to when $\tau=1$. Setting $\tau=0.9$ leads to the mining of approximately 15k objects, which yields the highest AP. This suggests that we have identified an ideal threshold value for mining the pseudo labels.
However, when $\tau$ is further reduced to 0.85, the number of mined objects increases explosively, and the model quickly collapses.
\subsubsection{Strategies For Using the Mined Lesions in Selective Retraining}
The mined lesion is ignored by the previous approach \cite{lyu2021segmentation}. We tested both ignore and adding strategies. The results are shown in Table \ref{tab:stra}. Both of these strategies bring substantial performance gain and adding the mined lesions to ground truth can give best result.

\section{Conclusion}
In this paper, we propose a novel explore training based ULD (ET-ULD) to address the challenge of universal lesion detection tasks with missing annotations. Our  approach focuses on recording the timing and frequency at which suspicious lesions are mined. To accomplish this, we introduce a prediction bank that stores all mined lesions during the training process. Visual and experimental results demonstrate that objects consistently appearing in the prediction bank are highly reliable true lesions, whereas objects that only emerge in the late stages of training are less reliable. By selectively choosing high-reliability mined lesions from the prediction bank for retraining, we significantly enhance the detection performance on both the DeepLesion and LiTS datasets, surpassing previous state-of-the-art results. Our approach provides valuable insights into addressing the missing annotation issue and offers promising advancements in lesion detection research.

\section*{Acknowledgments}


\bibliographystyle{IEEEtran}
\bibliography{tmi}

\end{document}